%% file: main.tex
\documentclass[sigconf, twocolumn, nonacm]{acmart}

\usepackage{todonotes}[false]
\usepackage{tabularx}
\usepackage{cleveref}
\usepackage{minted}
\usepackage{caption}
\usepackage{rotating}
\usepackage{adjustbox}
\usepackage{pdflscape}
\usepackage[table]{xcolor}
\usepackage{colortbl}  
\usepackage{array,collcell}
\usepackage{pgf} 

\newcommand{\framework}{\texttt{KGpipe}~}

\newcommand{\sspRDFa}{\ensuremath{SSP_{RDFa}}}
\newcommand{\sspRDFb}{\ensuremath{SSP_{RDFb}}}
\newcommand{\sspRDFc}{\ensuremath{SSP_{RDFc}}}

\newcommand{\sspJSONa}{\ensuremath{SSP_{JSONa}}}
\newcommand{\sspJSONb}{\ensuremath{SSP_{JSONb}}}
\newcommand{\sspJSONc}{\ensuremath{SSP_{JSONc}}}

\newcommand{\sspTEXTa}{\ensuremath{SSP_{TEXTa}}}
\newcommand{\sspTEXTb}{\ensuremath{SSP_{TEXTb}}}
\newcommand{\sspTEXTc}{\ensuremath{SSP_{TEXTc}}}

\newcommand{\exER}{\ensuremath{JSON_{ER}}}
\newcommand{\exKE}{\ensuremath{JSON_{KE}}}

\newcommand{\colorcell}[1]{%
  \edef\v{\fpeval{max(0,min(1,#1))}}%
  \edef\R{\fpeval{1-\v}}%
  \edef\G{\fpeval{\v}}%
  \edef\doCell{\noexpand\cellcolor[rgb]{\R,\G,0}}%
  \doCell{\v}%
}

\newcolumntype{G}{>{\collectcell\colorcell}c<{\endcollectcell}}

\newcommand{\colorcellf}[1]{%
  \edef\tmp{\fpeval{max(0.6,min(1,#1))}}%
  \edef\v{\fpeval{(\tmp - 0.6)/0.4}}%
  \edef\R{\fpeval{1-\v}}%
  \edef\G{\fpeval{\v}}%
  \edef\doCell{\noexpand\cellcolor[rgb]{\R,\G,0}}%
  \doCell{}#1%
}

\newcolumntype{F}{>{\collectcell\colorcellf}c<{\endcollectcell}}

\newif\ifshownotes
\shownotesfalse   

\begin{document}

\title{\framework: Generation and  Evaluation of Pipelines for Data Integration into Knowledge Graphs}

\author{Marvin Hofer}
\affiliation{%
  \institution{ScaDS.AI Dresden/Leipzig}
  \city{Leipzig}
  \state{Germany}
  \postcode{43017-6221}
}
\email{hofer@informatik.uni-leipzig.de}

\author{Erhard Rahm}
\orcid{0000-0002-2665-1114}
\affiliation{
  \institution{ScaDS.AI Dresden/Leipzig}
  \city{Leipzig}
  \country{Germany}
}
\email{rahm@uni-leipzig.de}

\begin{abstract}

Building high-quality knowledge graphs (KGs) from diverse sources requires combining methods for information extraction, data transformation, ontology mapping, entity matching, and data fusion. 
Numerous methods and tools exist for each of these tasks, but support for combining them into reproducible and effective end-to-end pipelines is still lacking.
We present a new framework, \framework for 
defining and executing integration pipelines that can combine existing tools or LLM (Large Language Model) functionality. 
To evaluate different pipelines and the resulting KGs, we propose a benchmark to integrate heterogeneous data of different formats (RDF, JSON, text) into a seed KG. 
We demonstrate the flexibility of \framework by running and comparatively evaluating several pipelines integrating sources of the same or different formats using selected performance and quality metrics.
\end{abstract}

\maketitle

\textbf{Artifact Availability:} \\
All code and datasets are available at: \\
\textbf{Repository:} \url{https://github.com/ScaDS/KGpipe} \\
\textbf{Dataset:} \url{https://doi.org/10.5281/zenodo.17246357} \\

\section{Introduction}

Knowledge graphs (KGs) have become essential for integrating and representing heterogeneous data in a unified and semantically rich form. Constructing KGs, however, remains complex, as it requires combining multiple specialized techniques, such as information extraction, ontology alignment, and entity matching. A huge amount of research and development in these areas has led to many powerful tools; some tasks are also well addressed by Large Language Models (LLMs)~\cite{DBLP:conf/aitomorrow/MeyerSFRJMDB023,freire2025large}.
However, these approaches are often developed independently, making it difficult to assemble them into reliable and reusable pipelines ~\cite{DBLP:journals/information/HoferOSKR24}.

In practice, KG pipelines are typically handcrafted for specific domains or tasks. This results in a proliferation of ad hoc scripts and configurations that are difficult to generalize or reproduce. Furthermore, the diversity of input data formats, ranging from RDF to JSON and unstructured text, adds to the integration challenge. There are a few tools to develop pipelines for constructing and updating KGs but these are mostly not openly available or  restricted to a single application and unable to reuse existing solutions for certain integration tasks ~\cite{DBLP:journals/information/HoferOSKR24}.
A further open challenge lies in evaluating KG pipelines because measuring and comparing the quality of an entire integration pipeline and the generated KGs is challenging and 
largely unsolved.  

To address these issues, we developed a framework \framework for building, executing, and evaluating comprehensive pipelines to integrate data sources of different formats (e.g., RDF, JSON, text) into a given KG. \framework is open-source and allows the automated composition of different task implementations within pipelines. The integration tasks have specified input and output formats to ensure that the output of one task can be consumed by the succeeding task.
Integration tasks can be executed as Python code, Docker containers, or remote HTTP services, hence supporting broad reusability and flexible integration across platforms.

We also propose and apply a new benchmark to comparatively evaluate different integration pipelines. The benchmark uses a reference KG and can evaluate the integration of overlapping sources of different formats into a seed KG using a set of metrics. We explore both the incremental integration of sources of the same format (single-source type integration, SSP) as well as of different formats (multiple-source type integration, MSP). 
This allows us to study the incremental evolution of the resulting KG for different pipelines and data formats, and specific aspects such as the impact of tasks like entity resolution or of the integration order of sources. 

This paper contributes:
\begin{enumerate}
    \item A new open-source framework \framework to define and execute  KG integration pipelines for different source formats by reusing and combining existing tools or LLM  functionality. 
    \item A new benchmark to evaluate KG integration pipelines providing datasets (seed KG, sources of different formats and a reference KG for comparison), a set of SSP and MSP integration pipelines for RDF, JSON, and text sources, and specifying a set of metrics to evaluate performance and quality of the pipelines. 
    \item We provide an initial evaluation demonstrating the applicability of \framework and providing insights into the effectiveness and efficiency of different pipelines and their task implementations. 
\end{enumerate}

In future work, 
we envision utilizing \framework to evaluate many pipelines for different settings to obtain training data for a largely automatic generation and configuration of pipelines.

After a discussion of related work we introduce the \framework framework and its implementation in Sections 3 and 4. Section 5 describes the \framework  benchmark which is applied in the evaluation in Section 6 before we conclude and provide an outlook to future work.

\section{Related Work}

Research on data integration for KGs spans from specialized tasks to models and  end-to-end systems for entire KG integration pipelines. 
\textbf{Individual integration tasks} are well studied and have resulted in numerous specialized toolsets and benchmark approaches. This holds particularly for information and relation extraction~\cite{DBLP:journals/eswa/Martinez-Rodriguez18/OpenIE}, schema/ontology matching~\cite{bellahsene2011schema,freire2025large},entity resolution (ER)~\cite{DBLP:journals/csur/ChristophidesEP21/ERSurvey} and KG error detection and knowledge completion ~\cite{DBLP:journals/semweb/Paulheim17/Refinement}. These tasks often involve already pipelines or workflows of several substeps, e.g. blocking, matching and clustering for entity resolution. 
Tools such as Magellan~\cite{DBLP:journals/cacm/DoanKCGPCMC20/Magellan} and JedAI~\cite{DBLP:conf/edbt/0001TTPSSIGPK20/JedAI} support full ER workflows, while Valentine~\cite{DBLP:conf/icde/KoutrasSIPBFLBK21/Valentine} provides an extensible execution and benchmarking system for schema matching. 

\textbf{Evaluations} of task-specific solutions and tools are typically conducted on static datasets, without considering incremental updates or interactions across integration steps. 
A recurring challenge across these task-specific solutions is pipeline configuration, i.e., selecting algorithms and tuning their parameters.  AutoER~\cite{nikoletos2024autoer/AutoER} formulates ER pipeline configuration as a search problem, using hyperparameter optimization or cross-dataset transfer learning. Hofer et al.~\cite{DBLP:conf/kgcw/HoferFR24/LLM4RML} demonstrate that large language models can generate RML~\cite{DBLP:conf/www/DimouSCVMW14RML/RML} mappings directly from semi-structured data, reducing manual mapping effort. While these works automate previously labor-intensive steps, they focus on single tasks, rather than the composition and coordination of multiple stages in an end-to-end integration pipeline.
Similarly, KG completion benchmarks (e.g., Open Graph Benchmark\footnote{\url{https://ogb.stanford.edu/docs/leader_linkprop/}}) rank models for link prediction or triple classification but do not capture upstream extraction quality or downstream fusion effects.

There is a growing number of surveys on approaches for \textbf{KG construction}, mostly describing the involved tasks as well as selected end-to-end solutions~\cite{weikum2021machine,hogan2021knowledge}.
Tamašauskaite and Groth~\cite{DBLP:journals/tosem/TamasauskaiteG23} review KG development processes and identify recurring lifecycle stages such as ontology design, extraction, processing, construction, and maintenance. 
In our recent survey 
~\cite{DBLP:journals/information/HoferOSKR24} we qualitatively evaluate available end-to-end systems such as NELL \cite{DBLP:conf/aaai/CarlsonBKSHM10/NELL}, SAGA ~\cite{DBLP:conf/sigmod/IlyasRKPQS22/SAGA}, XI ~\cite{cudre2020leveraging/Xi}, Plumber~\cite{DBLP:conf/icwe/Jaradeh0SBA21/Plumber} and Gawriljuk et al.~\cite{gawriljuk2016scalable},  against a set of requirements and note a lack of powerful open and modular tools to determine and execute KG construction pipelines that can reuse existing task-specific solutions. 
Furthermore, many previous systems evaluate only one handcrafted pipeline per domain, leaving open how alternative pipelines perform under controlled conditions and what influences the quality of the resulting KGs the most.  
This underlines the need of a benchmark to evaluate entire KP construction pipelines. 

Approaches for \textbf{quality validation} and monitoring complement these approaches by assessing generated KGs. 
Zaveri et al.~\cite{DBLP:journals/semweb/ZaveriRMPLA16/LDQual} provide a
classification of KG quality dimensions (e.g., accuracy, completeness, consistency, timeliness, accessibility, provenance) and associated metrics.
Sieve~\cite{DBLP:conf/edbt/MendesMB12/Sieve} introduces recency- and reputation-based scores to  asses the quality of linked data sources and uses these scores for resolving conflicts in data fusion.  
SHACL\footnote{\url{https://www.w3.org/TR/shacl/}} and RDFUnit~\cite{DBLP:conf/www/KontokostasWAHLCZ14/RDFUnit} support constraint checking, while PG-Schema~\cite{DBLP:journals/pacmmod/AnglesBD0GHLLMM23/PGSchema} provides schema guidance for property graphs. KGEval~\cite{DBLP:conf/emnlp/OjhaT17/KGEval} estimates KG correctness efficiently via constraint-based inference, and KGHeartBeat~\cite{DBLP:conf/esws/PellegrinoRT24/KGHeartbeat} offers a community-shared platform for longitudinal KG quality monitoring. 
These frameworks evaluate or enforce KG quality after construction, but do not benchmark how different integration pipelines lead to different quality outcomes.

Our work addresses the identified gaps and we (i) provide an open framework for composing multi-step integration pipelines from task-specific components, and (ii) release a benchmark for systematic comparative evaluation of entire KG integration pipelines. Rather than defining another single pipeline, \framework enables exploring the space of alternative pipelines and evaluating how their design choices influence KG quality across heterogeneous inputs.

\section{Knowledge Graph Integration}

A knowledge graph (KG) is a structured representation of information where entities (such as people, places, or products) are connected through relationships.
An ontology is a formal specification of concepts and the relationships between them within a domain, providing the shared vocabulary and logical rules that underpin knowledge graphs and other semantic systems.

In this work, we assume that KGs are represented in RDF. To limit the problem scope, we study data integration under a fixed KG ontology, leaving dynamic ontology changes for future work.
We refer to individual nodes in the knowledge graph as entities, and their assigned entity types correspond to ontology classes. Following common RDF terminology, entities have a unique human-readable name or \textit{label}, e.g., “Titanic (the movie)”. We use the term \textit{property} for any RDF predicate. Properties that link one entity to another are called relations, while properties that link an entity to a literal value are called attributes. Unless the distinction matters, we use property to refer to both relations and attributes. 

\subsection{Problem Definition}

The integration of heterogeneous sources into a knowledge graph (KG) can be realized through multiple alternative pipelines, each differing in task sequence, tool choice, and handling of specific source formats. Given a seed knowledge graph $KG_0$ defined under a target ontology $KGO$, and a sequence of input sources $S_1, S_2, \dots, S_n$ in various formats (e.g., RDF, JSON, text), the task is to construct a sequence of intermediate graphs 
\[
KG_1, KG_2, \dots, KG_n,
\] 
where each $KG_i$ results from integrating $S_i$ into $KG_{i-1}$.
A pipeline  \textit{p}   is formalized as a sequence of tasks:
\[
p = (T_1, T_2, \dots, T_k),
\]
where each task $T_i$ is defined by a function $F(I) = O$ with well-specified input formats $I$ and output formats $O$. 
A pipeline is valid if each task’s input is satisfied by either the initial input or the outputs of previous tasks, thus forming a correct chain of transformations that integrates one source into the KG.

Knowledge graph (KG) integration typically involves the following tasks:
\begin{itemize}
\item \textit{Information/Knowledge Extraction (KE):} transform raw input into a structured intermediate representation.
\item \textit{Data Mapping (DM):} project extracted data onto the target ontology or schema.
\item \textit{Schema Alignment/Ontology Mapping (SA/OM):} reconcile classes, relations, and attributes across datasets.
\item \textit{Entity Resolution (ER):} identify correspondences between new data and existing KG entities.
\item \textit{Entity Fusion (EF):} merge aligned entities and attributes, resolving conflicts and redundancy.
\item \textit{Data Cleaning (DC):} detect and correct errors or inconsistencies in source data or the KG.
\item \textit{Knowledge Completion (KC):} enrich the KG by inferring and adding missing entity types, attributes, and relations.
\end{itemize}
These tasks define the core integration workflow but can be specialized or extended depending on the input type and format. 

In addition to individual pipelines, we investigate the application of several pipelines to incrementally integrate multiple sources of either the same format or different formats  (single- vs. multi-source type pipelines, SSPs/MSPs). Integrating several sources is often needed in practice and brings additional challenges, such as possible accumulation of errors, increased likelihood of inconsistencies added to the KG, and dependencies on the order of integration. 

The central problem is that while many pipelines can be constructed, they are not equally effective. Different pipelines may yield widely varying results in terms of runtime, semantic correctness, coverage, and quality of the resulting KG. Thus, the task is not only to construct valid pipelines but to systematically evaluate, compare, and ultimately identify the best-performing pipelines for a given integration setting.

\subsection{Source-type-Aware Knowledge Graph Integration}
\label{sec:source-aware-kg-integration}

The tasks needed in a KG integration pipeline 
depend on the input format (structured, semi-structured, unstructured).
While the overarching task inventory is stable, not all tasks are necessary for every source, and format-specific structure (or lack thereof) strongly impacts the pipeline design. Pipelines are also not isolated: sources can be progressively transformed into more structured intermediate formats (e.g., Text $\rightarrow$ JSON $\rightarrow$ RDF), which allows use of a JSON (RDF) pipeline for text (JSON) input. 

\subsubsection{Structured Sources (RDF / Relational)}  
 
Here a typical pipeline is:
\emph{(i) Data Mapping (to RDF for relational input) $\rightarrow$ (ii) Schema Alignment $\rightarrow$ (iii) Entity Resolution $\rightarrow$ (iv) Entity Fusion (with provenance) $\rightarrow$ (v) Cleaning $\rightarrow$ (vi) Completion (optional).}

Although structured sources provide explicit schemas, integration is often complicated by divergent terminologies, overlapping entities, and conflicting attribute values. The main tasks in this pipeline are entity resolution, tightly coupled with ontology or schema matching. Reconciling identifiers, aligning vocabularies, and merging duplicates with provenance-aware fusion are also important.

\subsubsection{Semi-Structured Sources (JSON / XML / CSV)}

Typical Pipeline:
\emph{(i) Information Extraction (parse/normalize) $\rightarrow$ (ii) Data Mapping (to RDF) $\rightarrow$ (iii) Type Inference / Schema Alignment (as needed) $\rightarrow$ (iv) Entity Resolution $\rightarrow$ (v) Fusion $\rightarrow$ (vi) Cleaning.}

Semi-structured data introduces complexity through implicit or evolving schema, nested records, and heterogeneous field names. The central challenge here is schema interpretation and mapping. Converting JSON, XML, or CSV into graph-compatible form requires careful parsing and normalization, often combined with type inference to recover implicit semantics. Once this transformation is complete, subsequent resolution and fusion can build directly on techniques from the structured case.

\subsubsection{Unstructured Sources (Text / Web Pages / PDFs)}
Typical Pipeline:
\emph{(i) Knowledge Extraction (NER, relations, co-references) $\rightarrow$ (ii) Entity/Relation Linking $\rightarrow$ (iii) Schema Alignment (predicate normalisation) $\rightarrow$ (iv) Entity Resolution (duplicates) $\rightarrow$ (v) Cleaning $\rightarrow$ (vi) Completion.}

Unstructured sources such as text and web pages suffer from ambiguity, noise, and limited inherent structure. The main burden of integration falls on natural language processing, which transforms raw text into structured candidate triples. Named entity recognition, relation extraction, and co-reference resolution form the entry point, followed by linking entities and predicates to the target ontology. Because uncertainty at these tasks propagates downstream, knowledge extraction and linking dominate the pipeline. Frequently, intermediate representations such as JSON are produced, enabling the reuse of semi-structured or structured integration methods.

\paragraph{Overarching challenges and cross-cutting tasks}
Regardless of input format, integration is shaped by persistent challenges: the heterogeneity of schemas and vocabularies, the uncertainty inherent in matching or relation extraction, and the cumulative effects of local integration choices on overall KG quality. To counter these, pipelines routinely incorporate cleaning to detect and repair inconsistencies or duplicates, normalization to harmonize datatypes and surface forms, and completion to enrich the graph with missing links or types. These cross-cutting tasks stabilize the integration process across formats, ensuring that local steps contribute to global consistency and coverage. 

\section{Implementation}

\framework implements a modular and extensible architecture that enables the specification, validation, and execution of KG integration pipelines. Each pipeline consists of tasks that can be executed in multiple ways, with clear input/output specifications that ensure compatibility across steps.
To ensure reproducibility and easy debugging capabilities, our implementation uses file-based I/O between tasks. While this may introduce overhead compared to in-memory execution or direct streaming, it simplifies logging, inspection, and cross-language compatibility.
All tasks are executed sequentially within a pipeline. Parallelism and optimization are reserved for future work.

\subsection{Execution Modes}

To accommodate the diversity of tools and runtimes used in KG integration, our framework abstracts task execution from the underlying implementation. We support three execution backends for tasks:

\begin{itemize}
\item \textbf{Python}: Tasks are implemented as Python functions using libraries such as \texttt{rdflib}, \texttt{transformers}, or other domain-specific tools. 
\item \textbf{Docker}: Tasks packaged in containers can be executed in isolated environments, allowing the use of external tools or different language runtimes. This is useful for legacy code or language-specific dependencies.
\item \textbf{HTTP Services}: Remote tasks can be accessed via RESTful APIs. This allows scaling or externalizing complex tasks (e.g., large-scale entity linking or extraction models). 
\end{itemize}

These three backends are sufficient to cover the full spectrum of integration tasks we encounter: lightweight in-process operations (Python), encapsulated external tools or heterogeneous runtimes (Docker), and scalable remote services (HTTP). Supporting additional execution modes would largely duplicate functionality or add unnecessary complexity, whereas this set strikes a balance between flexibility, interoperability, and maintainability.

Pipelines are executed with a uniform interface, regardless of backend, and temporary files are used for communication between tasks.

\subsection{Task Interoperability}

A central challenge is ensuring that task outputs are compatible with subsequent task inputs, particularly across heterogeneous execution backends and data formats. For example, some entity matching tools produce CSV outputs while fusion components expect structured JSON; information extraction tools may output raw triples without ontology alignment; and linker components may require specific JSON schemas or XML.

To address this, we introduce intermediate transformation tasks that convert outputs into the formats expected by subsequent components. These exchange tasks act as glue between independently developed modules. Our interoperability strategy is inspired by Plumber~\cite{DBLP:conf/icwe/Jaradeh0SBA21/Plumber}, but it extends this approach to support other task areas.

Currently, we define two intermediate exchange formats:

\noindent\exER, used for entity resolution tasks, representing entity or relation matches as:
\begin{verbatim}
{ id1: $id # $number | $url
  id2: $id
  type: $str # "entity" | "relation"
  score: $float }
\end{verbatim}
\exKE, used for knowledge extraction tasks, captures text, triples, and links, the latter representing mappings of text forms to URIs in the KG. 
\begin{verbatim}
{ text: $str
  triples: [{head: $str, rel: $str, tail: $str}]
  links: [{form: $str, link: $id, score: $float}] } 
\end{verbatim}

\subsection{Validation and Constraints}

Each task specifies its expected input and output types, and during pipeline building, we apply static validation to ensure that consecutive tasks are compatible. This validation considers aspects such as file formats (e.g., TEXT, JSON, RDF), structural requirements like the presence of mandatory JSON keys or relation types, and, when available, schema compatibility between input and output. If a pipeline violates any of these constraints, it is rejected before execution.

To further support interoperability, these checks are aligned with the intermediate exchange formats introduced earlier (\exER and \exKE). By validating against these well-defined representations, we ensure that independently developed components can be safely composed into larger pipelines, even when they rely on different toolkits or programming languages.

\subsection{Single Source Pipelines}
\label{sec:pipelines}

\begin{figure}
    \centering
    \includegraphics[width=1\linewidth]{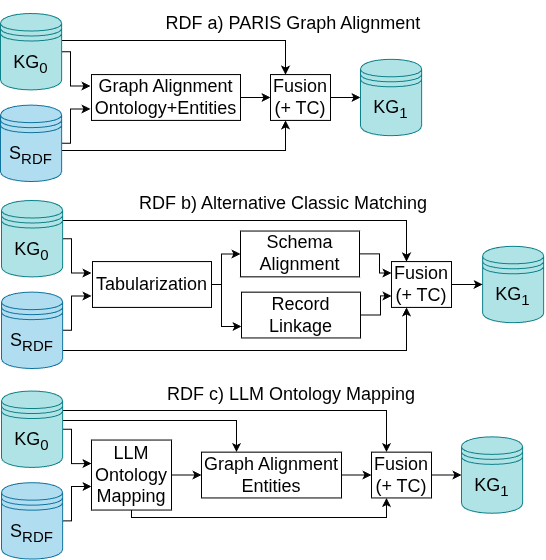}
    \caption{RDF single source pipeline layouts. TC=Type Completion.}
    \label{fig:ssp_rdf}
\end{figure}

\begin{figure}
    \centering
    \includegraphics[width=1\linewidth]{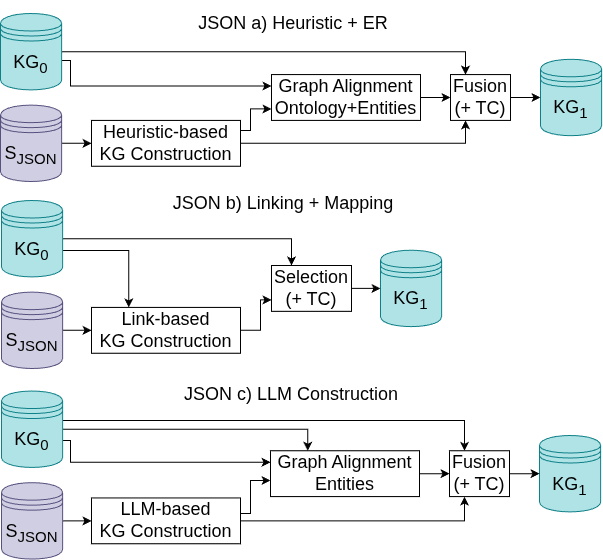}
    \caption{ JSON single source pipelines layouts. TC=Type Completion}
    \label{fig:ssp_json}
\end{figure}

\begin{figure}
    \centering
    \includegraphics[width=1\linewidth]{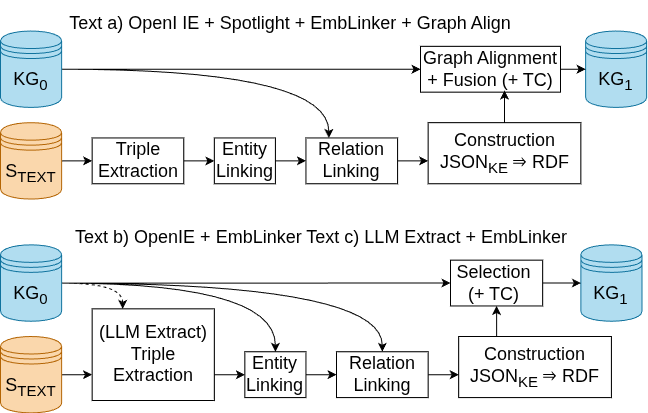}
    \caption{Text single source pipelines layouts. TC=Type Completion}
    \label{fig:ssp_text}
\end{figure}

\framework currently supports three source formats (RDF, JSON, TEXT) and we define three pipeline layouts per format, resulting in a total of nine SSP layouts (\Cref{fig:ssp_rdf,fig:ssp_json,fig:ssp_text}). The layouts follow a common pattern:
\begin{itemize}
\item an \emph{(a)} variant, representing a straightforward baseline pipeline,
\item a \emph{(b)} variant, which modifies or extends the baseline with alternative implementations, and
\item a \emph{(c)} variant, which uses a Large Language Model (LLM) to replace one or more tasks in the pipeline~\cite{freire2025large,DBLP:conf/vldb/ParciakVNPV24}.
\end{itemize}

Our goal is not to design the optimal pipeline for each source type, but to establish a set of solid and diverse pipelines that support a meaningful evaluation. We distinguish between a \emph{pipeline layout}, which specifies the sequence of tasks, and a \emph{pipeline configuration}, which fixes the tool or implementation used for each task.
\Cref{tab:tasks} compactly lists each implemented task used for our pipelines, with their name, task area, API/backend, and I/O contract.
Using the two exchange formats to keep tasks chainable: $JSON_{ER}$  for generated matches and $JSON_{KE}$ for the extracted triples and links. 
We reuse five external tools: PARIS~\cite{DBLP:journals/pvldb/SuchanekAS11/Paris} handles ontology/entity alignment over RDF (OM/EM), while JedAI~\cite{DBLP:conf/edbt/0001TTPSSIGPK20/JedAI} runs a simple record-linkage (EM) with a Valentine~\cite{DBLP:conf/icde/KoutrasSIPBFLBK21/Valentine} implementation performing schema matching (OM) on CSVs. For text tasks, Open Information Extraction (OpenIE)~\cite{DBLP:journals/eswa/Martinez-Rodriguez18/OpenIE} performs text triple extraction, and DBpedia Spotlight~\cite{DBLP:conf/i-semantics/MendesJGB11/Spotlight} recognizes and links entities to DBpedia identifiers.
Custom Python tasks include \textit{Tabularize} for RDF graph to CSV conversion, \textit{JSON-to-RDF} lifting JSON to a generic RDF based on a key-value to relation or property heuristic, \textit{JSON-Linking} that uses a similar heuristic but tries to directly tries to map constructed entities to the current KG, \textit{GenerateRDF}, to materialize the $JSON_{KE}$ data to RDF, and finally \textit{Fusion/SelectFirst} tasks which not only merge two RDF graphs (with resolved IDs from matches or by just selecting new entities and values) with a current KG first policy, but also performs \textit{type materialization} (adding the appropriate class type for the source/target entity of each fused triple via the ontology.
Three LLM-supported tasks: \textit{LLMExtract} for text triple-pattern extraction (TEXT$\rightarrow~JSON_{KE}$), \textit{LLMMapping} to construct ontology-compliant RDF from JSON given the ontology (JSON,RDF$\rightarrow$~RDF), and the RDF sample–based \textit{LLMMatcher} for ontology matching (RDF×2$\rightarrow~JSON_{ER}$).

\input{tables/tasks}

These tasks are the building blocks for the SSP layouts on RDF, JSON, and text source pipelines.
In the following, we describe the nine SSP layouts and configurations in more detail (see also \Cref{fig:ssp_rdf,fig:ssp_json,fig:ssp_text}).

\paragraph{RDF SSPs}
We implement three different RDF pipeline layouts, each with one configuration:

\begin{itemize}
    \item \sspRDFa~ applies a graph alignment method (PARIS~\cite{DBLP:journals/pvldb/SuchanekAS11/Paris}, executed inside Docker) to produce matches between entities in the seed and source RDF graphs. A fusion algorithm with a first-value preference is then applied: entity identifiers are resolved from source to seed, and for fusable relations, the first available value (typically from the source) is selected. Afterwards, type information on entities is inferred based on their current properties corresponding domain/range specs.
    \item \sspRDFb~ first transforms both RDF graphs into tabular CSV representations using a custom Python function. It then applies record linkage and schema matching to detect similar entities and relation names, using the clean–clean implementation of JedAi~\cite{DBLP:conf/edbt/0001TTPSSIGPK20/JedAI} and Valentine~\cite{DBLP:conf/icde/KoutrasSIPBFLBK21/Valentine} for matching. Finally, the fusion algorithm is applied as in \sspRDFa.
    \item \sspRDFc~ follows the same structure as \sspRDFa, but replaces the relation alignment to be generated by a Large Language Model (LLM). The LLM is used to match and map relations between the source and seed RDF graphs, based on sampled triples from each KG. PARIS is then applied for entity matching, followed by the same first-value fusion strategy.
\end{itemize}
 
\paragraph{JSON SSPs}
For the JSON format, we also have three layouts with a configuration each. 

\begin{itemize}
    \item \sspJSONa~ maps JSON data into a generic RDF graph, each key as a generic predicate URI, and types are constructed from the current path keys.
    The constructed RDF is then integrated into the seed KG following the same steps as in \sspRDFa.
    \item \sspJSONb~ uses a heuristic for label and property type detection (relation or attribute), it applies text embeddings to directly link JSON objects and keys to existing entities and relations in the given KG. If no label is found, a concatenation of object values is used to generate an entity embedding. Objects (entities) without a sufficiently similar match are assigned new entity identifiers.
    \item \sspJSONc~ avoids explicit mapping generation and instead prompts an LLM to directly output ontology-compliant RDF triples from the input JSON document. The LLM receives both the JSON data and the KG ontology as input and generates triples aligned with the ontology. Unlike approaches that require generating and selecting among candidate mappings, this strategy applies uniformly across all documents, thereby averaging out quality fluctuations in the LLM outputs over the dataset. This removes the need for manual or heuristic mapping selection, though it still introduces variability at the individual document level.
\end{itemize}

\paragraph{TEXT SSPs}
We use almost similar pipeline layouts for text sources, but with three different configurations:

\begin{itemize}
\item \sspTEXTa~ extracts triple patterns (entity surface forms) from the input text using OpenIE. Entities are linked with DBpedia Spotlight\footnote{DBpedia Spotlight is suitable for entity recognition and linking, because DBpedia widely covers the domain of interest.}, which maps mentions to the seed KG. Mentions not present in the seed are assigned new identifiers in a separate namespace. Relations are then linked using a custom embedding-based relation linker that maps text spans to ontology relations. The immediately constructed KG is integrated using the alignment and fusion tasks of \sspRDFa.
\item \sspTEXTb~ Also applies OpenIE, but for linking, it uses a text embedding model for both relation and entity linking (on labels or all values). Extracted entity mentions are compared with existing KG entities using text embeddings. As this approach links mentions directly to KG entities (ids), it bypasses the need for a separate entity matching or graph alignment step. The resulting triples are integrated into the KG, preserving existing values unless none are available.
\item \sspTEXTc~ replaces OpenIE with an LLM-based triple extraction component, which generates surface-form triples directly from the input text. Entity linking, relation linking, and fusion are performed as in \sspTEXTb.
\end{itemize}

For all LLM-based tasks, we use OpenAI’s \texttt{gpt-5-mini} model to balance quality, runtime, and cost.

\section{KG Pipeline Benchmark}

To evaluate integration pipelines, we propose a benchmark for a well-defined yet non-trivial domain: films and their related persons and companies. The movie domain offers a manageable scope while still presenting realistic integration challenges such as heterogeneous data formats, overlapping but incomplete entity coverage in source data, and schema-level heterogeneity. Extending the benchmark to additional domains is possible but time-consuming and left for future work.

We first describe the domain and KG ontology before providing details about the datasets (reference KG, seed KG, sources with overlapping entities) and evaluation metrics. 

\subsection{Domain and Ontology}

Our benchmark focuses on the \textbf{movie domain}, with entities such as films, actors, directors, and production companies. We manually curated a target ontology for this domain, which guides both pipeline integration and evaluation. The ontology comprises:

\begin{itemize}
    \item Three core classes: \texttt{Film}, \texttt{Person}, and \texttt{Company}, serving as the backbone of the domain model.  
    \item 25 properties (object relations or datatype properties, and counting the RDF \textit{label} and \textit{type} property), defined using OWL, RDF Schema, and SKOS, enabling consistent links across heterogeneous sources.  
    \item Annotations such as \texttt{rdfs:label} and \texttt{skos:altLabel} (to support lexical variation) and \texttt{owl:equivalentClass} or \\ \texttt{owl:equivalentProperty}) that facilitate schema alignment.  
    \item Integrity axioms, including \texttt{owl:disjointWith} (to catch invalid class overlaps, e.g., a \texttt{Film} typed as \texttt{Person}) and \texttt{owl:maxCardinality} (to follow rules, e.g., one \texttt{runtime} per film).  
\end{itemize}

\begin{figure}
    \centering
    \includegraphics[width=1\linewidth]{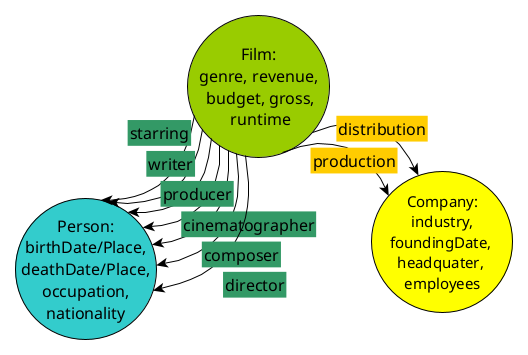}
    \caption{Ontology/Schema graph of classes film, person, company with their properties (relations and attributes).}
    \label{fig:ontology}
\end{figure}

A visualization of this graph schema (ontology) is shown in \Cref{fig:ontology}.  

A well-defined ontology is essential for the integration process because it provides the common target schema against which heterogeneous sources can be aligned. In the linking step, ontology classes and relations guide entity and schema matching, ensuring that different source fields map to the correct concepts. For cleaning, constraints and disjointness axioms allow the detection of errors such as a film mistakenly typed as a person or multiple runtimes assigned to the same movie. Finally, during completion, the ontology defines where additional values can be added consistently, for example, enriching a person entity with missing \texttt{birthPlace} or linking a film to its \texttt{director}. In short, the ontology acts as the semantic backbone that enables heterogeneous data to be reconciled into a coherent knowledge graph.

Although the ontology provides a clear target schema, integration remains challenging due to ambiguity and inconsistency in the sources. For example, some JSON infoboxes use the field name "date" without specifying whether it denotes a person’s birth or death date, which are distinguished in the ontology. 
Ambiguities also arise among role-related person or company properties for a film, e.g, \texttt{producer} person vs \texttt{production} company.
Properties may also use different data types. Runtime may appear in minutes, hours, or as free-text strings (“2h 15m”), and financial attributes such as \texttt{budget}, \texttt{gross}, and \texttt{revenue} are reported with inconsistent currencies and not always distinguished across sources.

\textit{Datasets} Our benchmark is built from a set of interrelated data artifacts that provide the basis for constructing and evaluating pipelines. Starting from a reference graph, we derive seed and source datasets in multiple formats (RDF, JSON, and text), each designed to expose specific integration challenges. These artifacts vary in coverage, structure, and representation, thereby reflecting the heterogeneity typical of real-world scenarios. 

\textit{Reference KG.} To establish a basis for evaluation, we first derive a reference by recursively collecting data for the film entities from DBpedia, and their related persons and companies, whenever such links are provided by the eight ontology relations.

We restrict our benchmark to 10,000 films rather than the full set of ~150,000 available in DBpedia (the final size is a bit lower due to cleaning and generating overlaps). This size already yields a knowledge graph of substantial complexity once related persons and companies are included, while keeping experiments computationally feasible and reproducible. It offers sufficient semantic variety to capture typical integration challenges without introducing unnecessary volume.

From this reference graph, we generate multiple entity splits to emulate realistic multi-source settings. Specifically, we partition the set of 10,000 films into four subsets of approximately 2,500 entities each. To introduce controlled redundancy, every pair of subsets (seed, sources) shares around 5\% of their film entities (30\% including persons and companies). This controlled overlap reflects the partial coverage commonly observed in real-world sources, where new sources usually cover existing entities in the KG.

\textit{Seed/Source type RDF.} The seed KG is defined as the first split of the reference graph. To generate the corresponding source RDFs, we create shaded versions of the remaining splits by renaming entity identifier namespaces. This ensures that the same real-world entities are represented with different IRIs across splits, thereby requiring entity resolution during integration.

\textit{Source type JSON.} For each film entity, we generate nested JSON records derived from 
one subgraph per film, including referenced person and company information.
Each record contains key–value pairs describing the film, such as title, actors, genre, or production company. 
JSON is a suitable source format for testing pipelines, since it is widely used on the Web, comes with less structural richness than RDF, and requires additional mapping to be aligned with the ontology.

\textit{Source type Text.} As a complementary unstructured source, we include DBpedia abstracts for all film entities. These textual descriptions provide narrative information mentioning films, people, and companies without explicit schema structure. Text is an important source type because it reflects the reality that much of the Web’s knowledge is available only in natural language and requires information extraction to be transformed into a knowledge graph.

\textit{Complementarity of Sources.} Together, RDF, JSON, and text capture the three principal degrees of structure encountered in real-world data: highly structured (RDF), semi-structured (JSON), and unstructured (text). This diversity ensures that the benchmark exercises pipelines across the full spectrum of integration challenges, from ontology alignment and schema mapping to entity resolution and information extraction.  

\textit{Supplementary Data.} In addition to the reference, seed, and source datasets, we provide supplementary data that supports evaluation and error analysis. First, we include metadata about the expected matches, i.e., the overlapping entities introduced during the dataset splitting. These records specify the entity type (e.g., \texttt{Film}) together with explicit links of the form \texttt{id1 = id2}, which serve as ground truth for entity resolution tasks. Second, we provide curated lists of verified source entities. These lists indicate which entities should appear in the integrated KG, enabling evaluators to check for missing or erroneously introduced entities. Together, these supplementary resources facilitate a more fine-grained assessment of pipeline performance beyond the structural and semantic properties of the resulting graphs.

We provide two smaller versions, with only a total of 100 and 1,000 film entities, for development and less resource-intensive benchmarks.  

\subsection{Evaluation Metrics}

Each pipeline is evaluated independently and compared against others using the following metrics.  
We group them into three complementary categories, each reflecting a distinct aspect of KG quality: \textbf{Statistical metrics} capture structural properties and resource usage, providing a baseline view of scale and coverage. \textbf{Semantic metrics} validate adherence to the target ontology, ensuring logical consistency and correctness beyond raw counts. \textbf{Reference metrics} measure fidelity against ground truth, either by comparison to a curated KG or by task-specific data. 
Together, these metrics evaluate structure, correctness, and truthfulness.

\subsubsection{Statistical Metrics}

We measure the structural characteristics of the generated KG:

\begin{itemize}
    \item $\mathbf{S_{FC}}$ \textbf{Fact Count}: Number of distinct triples (facts) in the KG.
    \item $\mathbf{S_{EC}}$ \textbf{Entity Count}: Number of distinct entities (nodes).
    \item $\mathbf{S_{RC}}$ \textbf{Relation Count}: Number of distinct relation names.
    \item $\mathbf{S_{TC}}$ \textbf{Type Count}: Number of distinct entity classes (types).
    \item $\mathbf{S_{UT}}$ \textbf{Count of untyped entities} that remain without a class, e.g., because none of their extracted or mapped properties match the ontology. 
    In such cases, type inference may not be able to assign a class leading to     isolated label-only entities.
    \item $\mathbf{S_D}$ \textbf{Graph Density}: Ratio of existing relations to the maximum possible number of relations between entities, indicating overall connectivity.  Identifying most relations results in a KG with high density while a KG with mostly isolated entities has low density.
\end{itemize}

\subsubsection{Resource Metrics}

These metrics capture the computational cost and efficiency of generating the KG (Quality-of-service).  

\begin{itemize}
    \item $\mathbf{Q_D}$ \textbf{Duration}: Total execution time of the pipeline in seconds.
    \item $\mathbf{Q_M}$ \textbf{Max Memory}: Peak memory consumption during runtime.
    \item $\mathbf{Q_C}$ \textbf{Additional Costs}: Non-computational overhead such as API usage, cloud hosting, or power consumption.
\end{itemize}

\subsubsection{Semantic Validation}

These metrics assess how well the generated KG adheres to the target ontology, ensuring logical consistency and correctness beyond raw counts (w.r.t. Ontology).
This is necessary as a corrupted semantic structure will influence the quality of later downstream tasks.

\begin{itemize}
    \item $\mathbf{O_{DT}}$ \textbf{Disjoint Types}: Number of entities assigned two or more mutually disjoint classes.  
    \emph{Example: an entity typed as both \texttt{Person} and \texttt{Company}, when these classes are declared disjoint.}
    \item $\mathbf{O_{D/R}}$ \textbf{Domain/Range}: Percentage of relations that respect their declared domain and range restrictions (affected by $\mathbf{O_{DT}}$).  
    \emph{Example: a relation \texttt{starring} must link an \texttt{Person} to a \texttt{Film}; a violation occurs if the subject is (an)other type than \texttt{Film}.}

    \item $\mathbf{O_{RD}}$ \textbf{Relation Direction}: Percentage of relations whose subject/object order is consistent with ontology definitions.  
    \emph{Example: \texttt{director} should link a \texttt{Film} (subject) to a \texttt{Person} (object); inverse direction counts as an error.}

    \item $\mathbf{O_{LT/F}}$ \textbf{Literal Datatype/Format}: Percentage of literal values consistent with their declared datatype and if the value has the correct format.  
    \emph{Example: a property \texttt{revenue} is of type double; values of type integer  or   string  are errors.}

\end{itemize}

\subsubsection{Reference Validation}
These metrics evaluate the fidelity of the generated KG against a manually curated reference KG~\cite{DBLP:journals/semweb/Paulheim17/Refinement} or task-specific verification data (like matches or links)~\cite{DBLP:journals/csur/ChristophidesEP21/ERSurvey}.  

\begin{itemize}
    \item $\mathbf{R_{EM}}$ \textbf{Entity Matching}: Precision and recall of expected vs.\ actual entity correspondences in entity resolution tasks.
    \item $\mathbf{R_{OM}}$ \textbf{Ontology Matching}: Precision and recall of expected vs.\ actual relation/property correspondences in ontology matching tasks.
    \item $\mathbf{R_{EL}}$ \textbf{Entity Linking}: Coverage of links between entities in the KG and external reference identifiers (e.g., ensuring each root entity is linked to at least one external reference).
    \item $\mathbf{R_{RL}}$ \textbf{Relation Linking}: Accuracy of links between input features (e.g., JSON keys) and ontology relations, where gold mappings are available. 
    \item $\mathbf{R_{SE}}$ \textbf{Source Entity Coverage}: Overlap between expected entities with their expected type (from a reference list, e.g., of entities in the source) and integrated entities.
    $\tilde{\mathbf{R}_{SE}}$: Same as $R_{SE}$ but allowing approximate string matching (e.g., label-based with high similarity).
    \item $\mathbf{R_{KG}}$ \textbf{Reference KG Coverage}: Exact triple overlap between the generated KG and a reference KG (subject, predicate, and object URIs match).
    $\tilde{\mathbf{R}}_{KG}$: overlap coverage is computed using a fuzzy entity alignment, while still requiring exact literal values, using matches on the entity labels.
    $\tilde{\mathbf{R}}'_{KG}$: Triple overlap with both fuzzy entity label matching and approximate literal value matching, to reference.
    Values are compared via semantic similarity (e.g., transformer embeddings).
\end{itemize}

We restrict the scope of some reference metrics to ensure valid and interpretable results.  
\textbf{Matching metrics} ($R_{EM}$, $R_{OM}$) are evaluated only for RDF integration pipelines.  
In this setting, entity identifiers can still be tracked reliably between the RDF graph and the initial source.  
For entities integrated from JSON or text, identifiers are often missing or inconsistent, which makes exact matching infeasible.
\textbf{Entity Linking} ($R_{EL}$) is evaluated only in terms of precision for \texttt{Film} entities in the seed KG and the current text split.  
Film entities provide a predictable anchor for evaluation, while links for other entity types are too sparse or unpredictable to measure recall meaningfully.

\input{tables/all_counts}
\begin{figure}[!ht]
    \centering
    \includegraphics[width=1\linewidth]{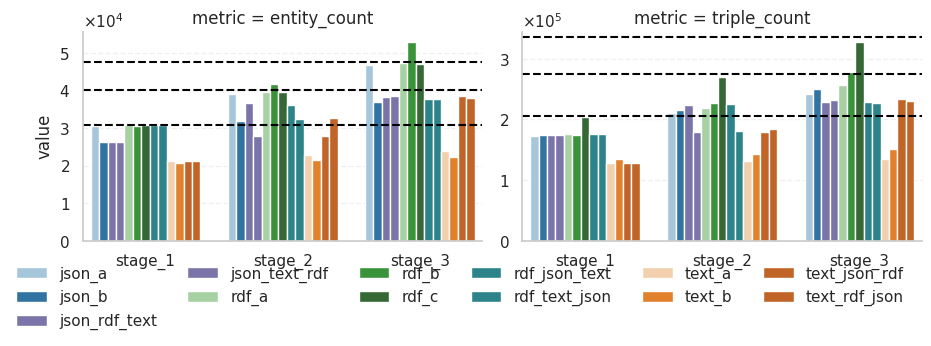}
    \caption{Visualization of statistical metrics (growth) for the 12 pipelines and their three increments/stages $KG_1-KG_3$. The black dotted lines indicate the expected reference KG sizes. All three SSP (c) pipelines are omitted here.}
    \label{fig:growth}
\end{figure}
\begin{figure*}[!ht]
    \centering
    \includegraphics[width=1\linewidth]{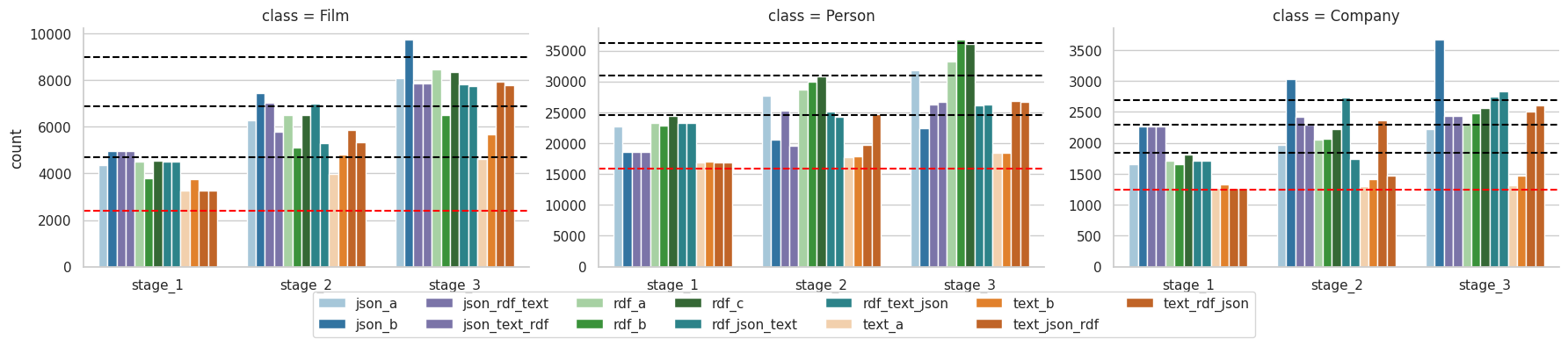}
    \caption{Compliance of integrated entities by type and expected entities at each source increment (stage) for all pipelines. The dotted lines indicate the values of the current reference between the seed (red) and the three source increments (black).} 
    \label{fig:entity_classes}
\end{figure*}
\input{tables/sem_scores}

\subsection{Aggregation and Ranking Method}
To better compare and rank different pipelines, it is desirable to aggregate the different metrics in a single score per pipeline. 
This can be based on a subset of the introduced metrics or perhaps additional ones. 
For our benchmark, we propose a two-step aggregation of metrics.
We first normalize and aggregate selected metrics per group and then determine a weighted average of the group metrics to obtain the overall aggregated score of a pipeline. 

Each group metric is the average of the  normalized metrics $Mn{_i}$ of the selected individual metrics $M{_i}$ per group: 
\[
GM_i = \frac{Mn_1+\dots+Mn_k}{k}.
\] 
We apply normalization to the metrics that are not yet within the range of 0 to 1. For the statistical metrics, we take the ratio of the different counts to the values of the reference graph. 
For the resource metrics duration and memory consumption, we relate the actual values of a pipeline to the minimal value $M_{min}$ of all pipelines:  
$$
Mn_{i}=\frac{M_{min}}{M_{i}} 
$$
For our evaluation, we apply four group  metrics: 
\begin{itemize}
\item 
    \textbf{$GM{_1}$ - Size score:} average of normalized  fact count and graph density.
\item 
    \textbf{$GM{_2}$ - Consistency score:} Average of violation scores (disjoint-domain, domain/range, direction, datatype/format).
\item 
    \textbf{$GM{_3}$ - Integration score:} Average of the reference KG coverage f-score, source entity-types integration score, plus the specific sources task ER/OM f-score (based on precision and recall values) or for text sources only film EL recall.
\item 
    \textbf{$GM{_4}$ - Efficiency score:} Normalized runtime duration and peak memory. This is the only score that directly depends on the used hardware environment. 
\end{itemize}

The total aggregated score $M_{\text{total}}(p)$  of a pipeline $p$ is the weighted average of the  group metrics:  
\[
M_{\text{total}}(p) = \sum_{i=1}^j w_i \cdot GM_i(p), \quad \sum_{i=1}^j w_i = 1,
\]

For the introduced four group metrics, we obtain: 
 \[
M_{\text{total}}= \alpha \cdot GM_{\text{1}} + \beta \cdot GM_{\text{2}} + \gamma \cdot GM_{\text{3}} + \delta \cdot GM_{\text{4}}
\]
where $\alpha, \beta, \gamma, \delta$ are weights that can be adjusted depending on the user’s goals (e.g., quality vs efficiency).

The weighting schemes applied for our evaluation are detailed in~\Cref{sec:ranking}.

\section{Evaluation}
Our evaluation mainly aims at demonstrating the usability and usefulness of the \framework framework and benchmark, but not for an exhaustive evaluation of the huge number of possible pipelines and their configuration. 
We focus on integrating three sources (data splits) into the seed KG by incrementally executing three pipelines to generate a final KG that can be compared to the reference KG. For the three input data formats, we apply nine single-source type pipelines (SSPs) when all splits have the same format and six multi-source type pipelines when each split is of a different format. For better readability, we adopt a more concise notation for pipeline references: for example, RDFa (or R\_A in tables) denotes the defined SSP RDFa, and RJT indicates the MSP of the SSP sequence RDF$\rightarrow$JSON$\rightarrow$Text.

All pipelines were executed on a machine equipped with an AMD Ryzen 9 9850HX processor (16 cores, 32 threads), 64 GB of main memory, a 1 TB SSD, and an NVIDIA RTX 4070 GPU. The availability of the GPU significantly increased the throughput of embedding-based task implementations. The LLM-based pipelines use OpenAI’s \texttt{gpt-5-mini}
with  API costs of RDFc=0.1€, JSONc=9.9€, and TEXTc=2.3€. 
The LLM-based pipelines for JSON and text had to be executed on the smaller 1k version of the benchmark to limit execution time and monetary expenses. 

To limit the number of MSP pipelines, we only consider cases where every split is of a different format and we only consider one of the a and b pipelines per step (RDFa, JSONb, TEXTa).
We apply fixed thresholds across all pipeline configurations to ensure consistency of evaluation and limit the number of pipeline executions: PARIS matches entities at a similarity threshold of 0.99 and relations at 0.5, JedAI matches at 0.5, Valentine matches at 0.1, LLM OM matches at 1.0, DBpedia Spotlight links at 0.8, and the embedding-based entity and relation linker (EmbEL/EmbRL) both at 0.8.

We evaluate the integration pipelines using the metrics introduced in the last section. Finally, we present different comparative rankings of the pipelines. 

\input{tables/erl_scores}
\input{tables/ref_scores}

\subsection{Size and Resource Statistics}

\Cref{tab:overview} shows the statistical and resource metrics for the SSP and MSP pipelines. 
\Cref{fig:growth,fig:entity_classes} illustrate how the KGs grow incrementally across the three integration stages. We discuss SSPs first, then MSPs, and finally relate them to the reference KG growth expectations.

While the relation/type counts are always correctly determined, tackled by the fusion implementation, there are significant differences in the other metrics (\Cref{tab:overview}). 
For the \textit{single-source type pipelines}, RDFb yields the largest KG in terms of entities, with more entities than in the reference graph (52.9k vs 47,7k). This is mainly due to a large number (7.5k) of untyped entities resulting from the CSV-based matching, which leads to many missed matches with properties of the ontology. The isolated entities could also not be matched and fused with other entities.  
RDFb also suffers from the highest cost of all a and b SSPs (6,491 s; 44.5 GB), mainly due to the expensive RDF$\rightarrow$CSV conversion plus  JedAI/Valentine matching.
The RDFa and JSONa pipelines achieve a similar number of entities as in the reference graph, but with only 76 / 72\% of the triples, indicating that many properties have not been integrated, as can also be seen from a substantial number of untyped entities.  By contrast, the LLM-based RDFc matches the size of the reference graph very well, with almost no untyped entities and their corresponding loss of properties. These advantages come at a relatively high runtime, though.
Despite very high runtimes, the text pipelines suffer from the biggest information loss, with only about half of the number of entities and less than 45\% of the triples of the reference graph. This is due to the inherent difficulty of correctly extracting entities and relations from unstructured text, as well as to possible limitations of the text extraction and linking tools. The LLM-based pipelines JSONc and TEXTc perform poorly and are only executed on the 1k seed KG. These pipelines were much slower, even on the 1k dataset, than other pipelines on the 10k version, indicating that LLM-based mapping to RDF is very slow.  

The six MSP pipelines perform very similarly and reach a medium number of entities that is higher than for the pure text pipelines but lower than for the RDF pipelines. This shows that the order of the source formats, e.g., whether text data is integrated first or last, does not influence the size of the final result KG. 

The development of the KG sizes across the three increments is shown in 
\Cref{fig:growth}. The pure text pipelines can hardly add further entities and triples compared to the seed graph. 
As expected, for the MSP pipelines, the text step generally adds the fewest entities and triples, while the RDF step adds the most.  We find similar trends in the development of the entity counts for the three entity types \texttt{Film}, \texttt{Person}, and \texttt{Company} (\cref{fig:entity_classes}).  Compared to the size of the reference graphs(s), the biggest gaps occur for the largest class \texttt{Person} except for the pure RDF pipelines. The JSONb pipeline determines more film and company entities than in the reference graph due to entities assigned to both classes, as a result of the type inference based on the wrongly detected entity and property mappings.

\subsection{Semantic and Reference Validation}

We first analyze the semantic correctness of the generated KGs according to the validation metrics defined in the last section.
\Cref{tab:semantic} shows that the RDF-based pipelines achieve almost perfect quality. The JSON pipelines are generally performing well despite some problems, especially for JSONb regarding the relation domain ($\mathbf{O_{D}}$) and range ($\mathbf{O_{R}}$). JSONa shows an issue with the literal datatypes ($\mathbf{O_{LT}}$), and JSONc has a few inversely mapped properties ($\mathbf{O_{RD}}$).
The TEXT pipelines introduce some relations with inverted direction (e.g., PERSON→FILM for starring).
TEXTb, TEXTc, and JSONb have the lowest structural correctness due to a larger number of entities assigned to more than one type ($\mathbf{O_{DT}}$) and erroneous relation domains and ranges. The six MSP pipelines perform rather similarly with an average correctness ratio of about 0.98. 

We now turn to the matching and linking quality of the different pipelines at different stages. 
 \Cref{tab:matching} shows the precision and recall results for entity and ontology (relation) matching, while \Cref{tab:linking} provides the linking quality (of film entities) for pipelines applying entity linking.
RDFa with the PARIS matcher achieves excellent entity matching quality, while the LLM-based RDFc performs best for ontology matching. RDFb has a reduced precision and recall of 0.81 to 0.89 due to limitations introduced by the conversion to CSV and tabular matcher. 
Film-level linking is generally of limited quality and especially poor for the text pipelines. TEXTa with DBpedia Spotlight is better here compared to the simpler methods based on keywords and embeddings in TEXTb and TEXTc. 
The MSP quality results vary greatly between stages, depending on whether matching or linking is performed. 

\Cref{tab:reference} summarizes, in terms of recall and precision, how effectively the pipelines integrate new entities from each source ($R_{SE}$) and how well they can reconstruct the reference KG ($R_{KG}$) across the three increments. 
The results align mostly with earlier observations 
especially regarding strengths of RDF and limitations for text sources. The precision values for source integration are not shown as they are always perfect since only data from the sources are integrated.  

Regarding source entity integration (left part of \Cref{tab:reference}), the RDFc pipeline achieves almost perfect recall (0.98), followed by RDFA and JSONa, with values around 0.9.
RDFb suffers from the poorer CSV-based matching and many wrongly fused entities. JSONb, which relies on embedding-based linking, misses almost half of the source entities.  
The text pipelines achieve by far the weakest source entity recall. This is consistent with low entity-linking recall in \Cref{tab:linking}, particularly for TEXTb and TEXTc.

For the overlap with the reference graph (right part of \Cref{tab:reference}), we focus on its portion without the seed graph - in contrast to the counts in \Cref{tab:overview}.  Again, the LLM-based RDFc pipeline performs best with very good recall and precision. The pipelines using entity linking (JSONb, TEXT pipelines) achieve a very low recall and therefore fail to correctly integrate most new entities. This is especially dramatic for the TEXT pipelines that can hardly add any new and correct entities to the seed graph.  
We observed, however, that there is no linear correlation between the quality of ontology matching and the resulting KG quality, e.g., overlap with the reference graph. This is because wrong match decisions for a frequently occurring property, e.g. \texttt{birthDate}, have a much stronger impact than that of less frequent properties, e.g. \texttt{deathDate}, although both properties have the same influence on the quality of ontology matching. 

For the MSP pipelines, the order in which sources are integrated leads to differences in the intermediate steps, with recall advantages for the pipelines starting with RDF (RJT, RTJ). However, after the three increments, all MSP pipelines are close together with a very good precision of 0.97-0.98 but a rather poor recall of only 0.31-0.33, mainly caused by the weak text step.   

Overall, the results confirm that strong entity/property matching and semantic consistency are critical for strong reference KG overlap, while weak entity linking, poor directionality, or type errors suppress overlap and harm incremental integration performance. Integrating structured (RDF) data has clear advantages over the integration of text data. The observed limitations show that the used pipelines and their configurations are far from perfect. This was expected, and the benchmark results indicate the biggest problems to tackle with better approaches and configurations (e.g., regarding entity and ontology matching, type inference, and integration of text data).

\input{tables/rank_scores}

\subsection{Comparative rankings}
\label{sec:ranking}

To compare pipelines holistically, we apply the aggregated pipelines based on the weighted average of the four group metrics (size score, consistency score, integration score, efficiency score). 

We compare the following five ranking options (weighting schemes):

\begin{itemize}
    \item 
    \textit{Equal weighting:} $(\alpha, \beta, \gamma, \delta) = (0.25, 0.25, 0.25, 0.25)$;
    \item 
    \textit{Quantity-oriented weighting:}  $(\alpha, \beta, \gamma, \delta) = (0.50, 0.10, 0.10, 0.30)$
    \item 
    \textit{Quality-focused weighting:}  $(\alpha, \beta, \gamma, \delta) = (0.0, 0.50, 0.50, 0.0)$
    \item 
    \textit{Reference-focused weighting:} $(\alpha, \beta, \gamma, \delta) = (0.0, 0.20, 0.8, 0.00)$
    \item \textit{Efficiency-oriented weighting:} $(\alpha, \beta, \gamma, \delta) = (0.20, 0.20, 0.20, 0.40)$
\end{itemize}

\Cref{tab:ranking} shows the calculated aggregated pipeline scores and five rankings, in the order top best to bottom worst.
Overall, RDFa and RDfc perform best.  RDFa achieves the most balanced performance, combining high structural correctness with good coverage and comparatively low runtime. The LLM-based RDFc is best regarding semantic quality and coverage of the reference KG, but at a higher computational cost. Among the JSON pipelines, JSONa ranks higher than JSONb due to a more stable structural behavior and improved semantic and reference quality. The text pipelines, especially TEXTb, consistently perform the weakest, largely due to limitations in extraction and linking.
All mixed-format pipelines fall into a middle range due to the limitations of the text processing step. The LLM pipelines JSONc and TEXTc are not directly comparable, as they are executed on the smaller dataset only and perform quite poorly.  

In summary, the results are quite mixed, indicating a high potential for improvement in most pipelines. It is hoped that the proposed framework and benchmark help to identify such improvements and provide insights for building effective data integration pipelines for KGs.

\section{Conclusions and outlook}

We introduced \framework, a modular and open framework for defining, executing, and evaluating integration pipelines to construct knowledge graphs from heterogeneous sources.
\framework supports the integration of structured and unstructured data by the combined use of existing tools as well as LLMs for critical tasks such as data transformation, text extraction, entity and relation linking, type completion, ontology, and entity matching and entity fusion. 
Furthermore, \framework supports the first benchmark to evaluate entire KG construction pipelines defined and applied in this paper. The benchmark from the movie domain includes a reference graph and evaluates pipelines to extend a seed KG by integrating several overlapping data sources of the same or different formats. The evaluation covers a wide spectrum of metrics that can be combined into a single score based on a configurable weighting scheme.  

To demonstrate the viability and potential of \framework we defined pipelines for different source formats (RDF, JSON, Text) and executed and evaluated them on the \framework benchmark. 
The evaluation allowed the comparative evaluation of the different pipelines and the tools used in them, as well as the identification of limitations to address within improved pipelines. 
We showed that the integration of RDF sources achieved better integration results compared to JSON and text, and that the text pipelines were especially weak in their current form.  The use of LLMs was only successful for RDF sources, but too slow and inefficient for JSON and text data.  The overall ranking resulted in the best performance and quality for the RDFa and RDFc pipelines. 

This work opens several directions for further development. 
First, we plan to expand the task library by adding further tasks such as data cleaning, knowledge completion, and ontology evolution. Second, the benchmark could be extended by an additional domain and further metrics.  
Third, we aim at a data-driven selection and tuning of pipelines. By accumulating pipeline–performance pairs across sources and integration settings, we aim to train models or LLM-based agents that can recommend pipeline designs and parameter configurations conditioned on a target ontology, seed graph, and source descriptors, optimizing multi-objective quality–cost trade-offs (size, consistency, integration performance, and efficiency) under explicit compute or latency budgets.

\begin{acks}
The authors acknowledge the financial support by the Federal Ministry of Education and Research of Germany and by the Sächsische Staatsministerium für Wissenschaft Kultur und Tourismus in the program Center of Excellence for AI-research "Center for Scalable Data Analytics and Artificial Intelligence Dresden/Leipzig", project identification number: ScaDS.AI.
\end{acks}

\bibliographystyle{ACM-Reference-Format}
\bibliography{research}

\end{document}

%% file: tables/tasks.tex
\begin{table}[!ht]
    \centering
    \begin{tabular}{lllcc}
\toprule
Name & Tasks & API & Input & Output \\ 
\midrule 
Paris     & EM, OM & D & RDFx2 & JSON$_{ER}$ \\
Valentine & OM & D & CSVx2 & JSON$_{ER}$ \\
JedAI     & EM & D & CSVx2 & JSON$_{ER}$ \\
StanfordOpenIE    & TE & D & TEXT & JSON$_{KE}$ \\
Spotlight & EL & D/H & TEXT & JSON$_{KE}$ \\
\midrule
EmbeddingEL & EL & Py & JSON$_{KE}$ & JSON$_{KE}$ \\
EmbeddingRL & RL & Py & JSON$_{KE}$ & JSON$_{KE}$ \\
Json-Linking & EL, RL & Py & JSON & JSON$_{KE}$ \\
Tabularize     & DM & Py & RDF & CSV \\
Json-to-RDF    & DM & Py & JSON & RDF \\ 
GenerateRDF$_{KE}$     & DM & Py & JSON$_{KE}$ & RDF \\
FusionFirst     & EF, TC & Py & RDFx2, JSON$_{ER}$ & RDF \\
SelectFirst     & EF, TC & Py & RDFx2 & RDF \\
\midrule
LLMExtract & TE & H & TEXT & JSON$_{KE}$ \\
LLMMapping & DM & H & JSON, RDF & RDF \\
LLMMatcher & OM & H & RDFx2 & JSON$_{ER}$ \\
\bottomrule
    \end{tabular}
    \caption{Current Integration Tasks of \framework: OM=Ontology Matching, EF=Entity Fusion, EM=Entity Matching, OM=Ontology Matching, TE=Text Extraction, EL=Entity Linking, RL=Relation Linking, DM=Data Mapping, TC=Type Completion. APIs: D=Docker, H=Http, Py=Python.} 
    \label{tab:tasks}
\end{table}

%% file: tables/all_counts.tex
\begin{table}
\centering
\begin{tabular}{lrrrrrrr}
\toprule
Pipeline  & FC      & EC     & RC & TC & UT & D (s) & M (GB)  \\
\midrule
Seed 1k & 16,417 & 2,793 & 25 & 3 & - & - & - \\
Seed 10k & 123,686 & 19,527 & 25 & 3 & - & - & - \\
Ref. 1k$^*$   & 63,359  & 8,935  & 25 & 3  & -  & -   & -      \\
Ref. 10k  & 336,002 & 47,706 & 25 & 3  & -   & -  & -     \\
\midrule
J\_A     & 242,319 & 46,657 & 25 & 3  & 4,801    & 77    & 4.3  \\
J\_B     & 249,547 & 36,977 & 25 & 3  & 3,359    & 166   & 5.6  \\
J\_C*     & 49,787  & 7,668  & 25 & 3  & 48       & 7,025 & 4.8  \\
R\_A     & 256,528 & 47,258 & 25 & 3  & 3,500    & 65    & 6.3  \\
R\_B     & 278,366 & 52,932 & 25 & 3  & 7,512    & 6,491 & 44.5 \\
R\_C     & 327,714 & 47,013 & 25 & 3  & 273     & 222   & 6.1  \\
T\_A     & 133,883 & 23,946 & 25 & 3  & 72      & 1,004 & 19.8 \\
T\_B     & 150,541 & 22,185 & 25 & 3  & 33      & 1,030 & 10.4 \\
T\_C*     & 24,134  & 3,608  & 25 & 3  & 3       & 1,963 & 10.4 \\
JRT      & 227,846 & 38,375 & 25 & 3  & 2,960    & 411   & 19.8 \\
JTR      & 231,706 & 38,541 & 25 & 3  & 2,727    & 398   & 19.8 \\
RJT      & 228,398 & 37,654 & 25 & 3  & 2,279    & 429   & 19.8 \\
RTJ      & 227,774 & 37,745 & 25 & 3  & 2,308    & 408   & 19.8 \\
TJR      & 232,989 & 38,619 & 25 & 3  & 2,655    & 411   & 19.8 \\
TRJ      & 230,945 & 37,980 & 25 & 3  & 2,327    & 417   & 19.8 
\\ \bottomrule
\end{tabular}
\caption{Pipeline result statistics: FC=Fact Count, EC=Entity Count, RC=Relation Count, TC=Type Count, UT=Un-Typed Entities, D=Duration in seconds, M=Memory in gigabyte. The (C)$^*$ pipelines are only executed for the 1k version of the benchmark. }
\label{tab:overview}
\end{table}

%% file: tables/sem_scores.tex
\begin{table}
\centering
\begin{tabular}{lccccccF}
\toprule
                 & $O_{DT}$    & $O_{D}$                & $O_{R}$               & $O_{RD}$                  & $O_{LT}$       & $O_{LF}$               & \multicolumn{1}{c}{$O_{Avg}$}  \\
\midrule
J\_A & 0.995 & 0.996 & 0.842 & 1     & 0.794 & 1     & 0.938 \\
J\_B & 0.941 & 0.933 & 0.931 & 1     & 1     & 1     & 0.968 \\
J\_C* & 0.987 & 0.987 & 0.992 & 0.992 & 1     & 1     & 0.993 \\
R\_A & 0.994 & 0.995 & 0.988 & 1     & 1     & 1     & 0.996 \\
R\_B & 0.992 & 0.993 & 0.974 & 1     & 1     & 1     & 0.993 \\
R\_C & 0.993 & 0.996 & 0.989 & 1     & 1     & 1     & 0.996 \\
T\_A & 0.978 & 0.969 & 0.974 & 0.997 & 1     & 0.997 & 0.986 \\
T\_B & 0.855 & 0.725 & 0.881 & 0.911 & 1     & 0.985 & 0.893 \\
T\_C* & 0.737 & 0.551 & 0.867 & 0.849 & 1     & 0.939 & 0.824 \\
\midrule
JRT  & 0.97  & 0.961 & 0.962 & 0.999 & 1     & 1     & 0.982 \\
JTR  & 0.97  & 0.961 & 0.958 & 0.999 & 1     & 0.999 & 0.981 \\
RJT  & 0.966 & 0.961 & 0.959 & 0.999 & 1     & 1     & 0.981 \\
RTJ  & 0.964 & 0.957 & 0.952 & 0.999 & 1     & 0.999 & 0.979 \\
TJR  & 0.966 & 0.958 & 0.957 & 0.999 & 1     & 0.999 & 0.980 \\
TRJ  & 0.964 & 0.959 & 0.957 & 0.999 & 1     & 0.999 & 0.980 
\\ \bottomrule
\end{tabular}
\caption{Semantic Structure Scores.}
\label{tab:semantic}
\end{table}

%% file: tables/erl_scores.tex
\begin{table}
\parbox{.45\linewidth}{
\centering
\setlength{\tabcolsep}{3pt}
\begin{tabular}{llGGGG}
\toprule
Pipe    & I & \multicolumn{1}{c}{$EM_{p}$} & \multicolumn{1}{c}{$EM_{r}$} & 
\multicolumn{1}{c}{$OM_{p}$} &  \multicolumn{1}{c}{$OM_{r}$} \\
\midrule
J\_A & 1 & 0.99 & 0.41 & 0.43 & 0.38 \\
J\_A & 2 & 0.75 & 0.40 & 0.43 & 0.38 \\
J\_A & 3 & 0.66 & 0.40 & 0.43 & 0.38 \\
R\_A & 1 & 0.98 & 1.00 & 1.00 & 0.63 \\
R\_A & 2 & 0.98 & 1.00 & 1.00 & 0.58 \\
R\_A & 3 & 0.98 & 1.00 & 1.00 & 0.63 \\
R\_B & 1 & 0.83 & 0.88 & 0.89 & 0.73 \\
R\_B & 2 & 0.81 & 0.88 & 0.89 & 0.77 \\
R\_B & 3 & 0.82 & 0.88 & 0.89 & 0.77 \\
R\_C & 1 & 0.98 & 1.00 & 1.00 & 0.83 \\
R\_C & 2 & 0.98 & 1.00 & 1.00 & 0.83 \\
R\_C & 3 & 0.98 & 1.00 & 1.00 & 0.75 \\
\midrule
JRT  & 2 & 0.97 & 1.00 & 1.00 & 0.58 \\
JTR  & 3 & 0.97 & 1.00 & 1.00 & 0.58 \\
R\_\_  & 1 & 0.98 & 1.00 & 1.00 & 0.63 \\
TJR  & 3 & 0.97 & 1.00 & 1.00 & 0.63 \\
TRJ  & 2 & 0.98 & 1.00 & 1.00 & 0.58 
\\ \bottomrule
\end{tabular}
\caption{Matching scores over all entity types.}
\label{tab:matching}
}
\hfill
\parbox{.45\linewidth}{
\centering
\begin{tabular}{llG}
\toprule
Pipeline  & inc & \multicolumn{1}{c}{$EL_R$}    \\
\midrule
J\_B & 1 & 0.64 \\
J\_B & 2 & 0.62 \\
J\_B & 3 & 0.62 \\
T\_A & 1 & 0.43 \\
T\_A & 2 & 0.43 \\
T\_A & 3 & 0.45 \\
T\_B & 1 & 0.13 \\
T\_B & 2 & 0.12 \\
T\_B & 3 & 0.12 \\
T\_C* & 1 & 0.21 \\
T\_C* & 2 & 0.14 \\
T\_C* & 3 & 0.16 \\
\midrule
JTR  & 2 & 0.43 \\
JRT  & 3 & 0.45 \\
RTJ  & 2 & 0.43 \\
RJT  & 3 & 0.45 \\
T\_\_  & 1 & 0.43 
\\ \bottomrule
\end{tabular}
\caption{Linking scores for film entities over text doc.}
\label{tab:linking}
}
\end{table}

%% file: tables/ref_scores.tex
\begin{table}
\centering
\setlength{\tabcolsep}{3pt}
\begin{tabular}{l|GGG|GcGcGc}
\toprule
& \multicolumn{3}{c|}{$R_{SE}$ @inc} & \multicolumn{6}{c}{$R_{KG}$ @inc (no Seed)}  \\
Pipe       & 
\multicolumn{1}{c}{r@1} &  
\multicolumn{1}{c}{r@2} &  
\multicolumn{1}{c|}{r@3} &

\multicolumn{1}{c}{p@1} &
\multicolumn{1}{c}{r@1} &  
\multicolumn{1}{c}{p@2} &
\multicolumn{1}{c}{r@2} &  
\multicolumn{1}{c}{p@3} &
\multicolumn{1}{c}{r@3} \\
\midrule
J\_A & 0.88 & 0.88 & 0.88 & 0.98 & 0.58 & 0.98 & 0.56 & 0.98 & 0.54 \\
J\_B & 0.55 & 0.53 & 0.52 & 0.98 & 0.21 & 0.97 & 0.21 & 0.96 & 0.2  \\
J\_C* & 0.73 & 0.74 & 0.7  & 0.78 & 0.57 & 0.77 & 0.57 & 0.75 & 0.55 \\
R\_A & 0.91 & 0.91 & 0.93 & 1    & 0.64 & 0.99 & 0.61 & 0.99 & 0.61 \\
R\_B & 0.83 & 0.83 & 0.83 & 0.99 & 0.56 & 0.98 & 0.54 & 0.98 & 0.52 \\
R\_C & 0.98 & 0.98 & 0.97 & 1    & 0.95 & 0.99 & 0.93 & 0.98 & 0.92 \\
T\_A & 0.4  & 0.39 & 0.39 & 0.52 & 0.01 & 0.52 & 0.01 & 0.53 & 0.01 \\
T\_B & 0.38 & 0.36 & 0.35 & 0.62 & 0.01 & 0.57 & 0.01 & 0.56 & 0.01 \\
T\_C* & 0.17 & 0.17 & 0.16 & 0.37 & 0.02 & 0.38 & 0.02 & 0.37 & 0.02 \\
JRT  & 0.55 & 0.87 & 0.51 & 0.98 & 0.21 & 0.99 & 0.43 & 0.98 & 0.31 \\
JTR  & 0.55 & 0.41 & 0.88 & 0.98 & 0.21 & 0.95 & 0.12 & 0.98 & 0.32 \\
RJT  & 0.92 & 0.61 & 0.5  & 1    & 0.64 & 0.98 & 0.43 & 0.97 & 0.31 \\
RTJ  & 0.91 & 0.51 & 0.6  & 1    & 0.64 & 0.98 & 0.35 & 0.97 & 0.31 \\
TJR  & 0.41 & 0.54 & 0.89 & 0.52 & 0.01 & 0.95 & 0.12 & 0.98 & 0.33 \\
TRJ  & 0.41 & 0.91 & 0.6  & 0.52 & 0.01 & 0.98 & 0.37 & 0.97 & 0.32 
\\ \bottomrule
\end{tabular}
\caption{Source entity integration recall (left side) and overlap with the reference graph (right side), without checking seed triples, and using relaxed IDs+values,  after each increment (new sources 1–3).}
\label{tab:reference}
\end{table}

%% file: tables/rank_scores.tex
\begin{table}[ht]
\centering
\setlength{\tabcolsep}{3pt}
\begin{tabular}{lGlGlGlGlG}
\toprule
eq. & \multicolumn{1}{c|}{rank}  & quan. & \multicolumn{1}{c|}{rank} & qual. & \multicolumn{1}{c|}{rank} & ref. & \multicolumn{1}{c|}{rank}  & eff. & \multicolumn{1}{c}{rank}   \\
\midrule
R\_A & 0.902 & R\_A & 0.866 & R\_C & 0.977 & R\_C & 0.966 & R\_A & 0.922 \\
R\_C & 0.806 & J\_A & 0.778 & R\_A & 0.924 & R\_A & 0.88  & J\_A & 0.808 \\
J\_A & 0.801 & R\_C & 0.771 & R\_B & 0.888 & R\_B & 0.825 & R\_C & 0.703 \\
J\_B & 0.676 & J\_B & 0.687 & J\_C* & 0.832 & J\_A & 0.751 & J\_B & 0.618 \\
R\_B & 0.642 & J\_C* & 0.591 & J\_A & 0.821 & J\_C* & 0.735 & R\_B & 0.516 \\
J\_C* & 0.629 & R\_B & 0.572 & J\_B & 0.733 & J\_B & 0.592 & J\_C* & 0.505 \\
RJT  & 0.58  & TJR  & 0.571 & RJT  & 0.702 & RJT  & 0.534 & RTJ  & 0.495 \\
RTJ  & 0.58  & JTR  & 0.571 & RTJ  & 0.698 & RTJ  & 0.53  & JTR  & 0.495 \\
TRJ  & 0.579 & TRJ  & 0.571 & TRJ  & 0.696 & TRJ  & 0.525 & TJR  & 0.494 \\
TJR  & 0.578 & RJT  & 0.569 & JRT  & 0.694 & JRT  & 0.522 & RJT  & 0.494 \\
JTR  & 0.578 & RTJ  & 0.569 & TJR  & 0.692 & TJR  & 0.52  & TRJ  & 0.494 \\
JRT  & 0.576 & JRT  & 0.564 & JTR  & 0.692 & JTR  & 0.518 & JRT  & 0.492 \\
T\_A & 0.482 & T\_B & 0.476 & T\_A & 0.634 & T\_A & 0.424 & T\_A & 0.398 \\
T\_B & 0.456 & T\_A & 0.443 & T\_B & 0.528 & T\_B & 0.31  & T\_B & 0.377 \\
T\_C* & 0.412 & T\_C* & 0.436 & T\_C* & 0.477 & T\_C* & 0.269 & T\_C* & 0.336 
\\ \bottomrule
\end{tabular}
\caption{Pipelines with different rankings, from best at the top to worst at the bottom, for five different weighting scenarios: \textbf{eq}al, \textbf{quan}tity, \textbf{qual}ity, \textbf{ref}erence, and \textbf{eff}iciency.}
\label{tab:ranking}
\end{table}